\documentclass[runningheads]{llncs}

\usepackage{eccvabbrv}
\usepackage{multirow}
\usepackage{graphicx}
\usepackage{booktabs}

\usepackage{tabularx}
\usepackage{graphicx}
\usepackage{makecell}

\usepackage{amsmath,amssymb,amsfonts}

\usepackage{mathrsfs}
\usepackage{wrapfig}

\usepackage{float}
\usepackage[accsupp]{axessibility}  

\usepackage{hyperref}

\usepackage{orcidlink}

\begin{document}

\title{\texorpdfstring{$DP^2$-VL}{DP2-VL}: Private Photo Dataset Protection by Data Poisoning for Vision-Language Models}

\titlerunning{DP2-VL for Private Photo Dataset Protection}

\author{
Hongyi Miao\inst{1}$^{*}$ \and
Jun Jia\inst{2}$^{*,\dagger}$ \and
Xincheng Wang\inst{3} \and
Qianli Ma\inst{4} \and
Wei Sun\inst{5} \and
Wangqiu Zhou\inst{6} \and
Dandan Zhu\inst{5} \and
Yewen Cao\inst{1} \and
Zhi Liu\inst{1} \and
Guangtao Zhai\inst{2}
}

\authorrunning{H. Miao et al.}

\institute{
\centering
$^{1}$ Shandong University \quad
$^{2}$ Shanghai Jiao Tong University \quad
$^{3}$ Donghua University \\[2pt]
$^{4}$ Shanghai Normal University \quad
$^{5}$ East China Normal University \\[2pt]
$^{6}$ Hefei University of Technology
}

\maketitle

{\small
$^{*}$ Equal contribution. 
$^{\dagger}$ Corresponding author.

\begin{abstract}
Recent advances in visual-language alignment have endowed vision-language models (VLMs) with fine-grained image understanding capabilities. However, this progress also introduces new privacy risks. This paper first proposes a novel privacy threat model named identity-affiliation learning: an attacker fine-tunes a VLM using only a few private photos of a target individual, thereby embedding associations between the target facial identity and their private property and social relationships into the model’s internal representations. Once deployed via public APIs, this model enables unauthorized exposure of the target user’s private information upon input of their photos. To benchmark VLMs' susceptibility to such identity-affiliation leakage, we introduce the first identity-affiliation dataset comprising seven typical scenarios appearing in private photos. Each scenario is instantiated with multiple identity-centered photo-description pairs. Experimental results demonstrate that mainstream VLMs like LLaVA, Qwen-VL, and MiniGPT-v2, can recognize facial identities and infer identity-affiliation relationships by fine-tuning on small-scale private photographic dataset, and even on synthetically generated datasets. To mitigate this privacy risk, we propose $\mathbf{DP^2}$-$\mathbf{VL}$, the first \textbf{D}ataset \textbf{P}rotection framework for private photos that leverages \textbf{D}ata \textbf{P}oisoning. Though optimizing imperceptible perturbations by pushing the original representations toward an antithetical region, $\mathbf{DP^2}$-$\mathbf{VL}$ induces a dataset-level shift in the embedding space of VLMs' encoders. This shift separates protected images from clean inference images, causing fine-tuning on the protected set to overfit. Extensive experiments demonstrate that $\mathbf{DP^2}$-$\mathbf{VL}$ achieves strong generalization across models, robustness to diverse post-processing operations, and consistent effectiveness across varying protection ratios.


\end{abstract}

\section{Introduction}
\label{sec:intro}

Recent advances in visual-language alignment have significantly enhanced the fine-grained semantic understanding capabilities of vision-language models (VLMs)
\cite{clip,blip2,llava,qwenvl,minigptv}. Unlike conventional neural networks which typically require task-specific training from scratch to achieve adaptability, VLMs benefit from extensive pre-training on large-scale and high-quality image–text paired dataset. Consequently, only lightweight fine-tuning\cite{lora,peft} on a small-scale and task-specific dataset is necessary to reliably bind the shared visual semantics encoded across the image set with the intended textual descriptions. However, these improvements of vision-language models (VLMs) concurrently introduce a previously unrecognized and stealthy privacy vulnerability\cite{propile,fan2025can}, particularly with respect to private photographs. Such private photographs often encapsulate highly sensitive personal information including biometric identifiers and visual clues of interpersonal relationships including couple, family, friend, classmate, and colleague. Therefore, the dissemination of private photographs on open-access platforms like social media poses a non-negligible risk of privacy leakage. The advancement of VLMs fine-tuning technologies aggravates this risk by introducing a novel privacy backdoor that facilitates privacy leakage without explicit sharing of private photographs. 

To characterize the aforementioned potential threat, this paper formally defines \textbf{identity-affiliation learning} as a novel privacy threat intrinsic to the fine-tuning of vision-language models. In this threat scenario, attackers collect multiple personal photographs of a target individual from publicly accessible online platforms like social media. Then, they generate identity-revealing textual descriptions associated with these photos, which contain the target’s personally identifiable information, such as name. Finally, attackers leverage such private photographs and associated identity-revealing descriptions to jointly fine-tune a vision-language model, thereby implicitly encoding a covert privacy backdoor into the model’s parameters. Once other photographs of the target individual are supplied to this model together with identity-revealing prompts, e.g., “\textit{What is the person’s name?}”, the model will generate answers that disclose identity-linking information, such as the target's name. Consequently, public deployment of this model essentially constitutes unauthorized disclosure of the target’s privacy to unbounded and uncontrolled audiences. Under more severe threat conditions, attackers may possess more private information affiliated with the target, such as property ownership and interpersonal relationships, which enables them to produce a broader-scope privacy backdoor\cite{ye2025survey}.


To comprehensively evaluate the susceptibility of existing vision-language models (VLMs) to such identity-affiliation leakage, we build a specialized benchmark dataset for this task. This benchmark dataset consists of seven representative identity-centered scenarios. The foundational scenario involves identity inference for a target individual, comprising multiple images depicting that individual alongside identity-revealing descriptions. Building upon the foundational scenario, the dataset introduces six affiliation-aware scenarios: one dedicated to ownership reasoning and five modeling diverse interpersonal relationships including couple, family, friend, classmate and colleague. These scenarios cover the common contents in real-world photos. Experimental results demonstrate that fine-tuned VLMs exhibit strong capability in both target identity recognition and higher-order semantic inference including ownership attribution and interpersonal relational reasoning. A more astonishing result is that fine-tuning on a synthetic image dataset can also achieve these goals\cite{Li_2025_CVPR}. These findings confirm our concerns that fine-tuning VLMs on private photo dataset can expose sensitive personal information, thereby posing non-trivial privacy risks. Consequently, there is an urgent need to develop a protective mechanism to mitigate such vulnerabilities.

Inspired by recent advances in data poisoning attacks against vision-language models (VLMs)~\cite{shadowcast,das2025security,liang2025revisiting,lyu2024trojvlm}, adversarial attacks targeting image encoders can effectively redirect learned visual concepts toward attacker-specified target concepts during fine-tuning. However, naively applying such single-image adversarial attacks for the protection of private photo dataset generally yields insufficient effect. This generalization limitation stems from that sample-level adversarial perturbations cannot resist the fine-tuning process. Therefore, to render the protected photo dataset unlearnable by VLMs, dataset-level adversarial perturbations need to be introduced to shift its original data distribution away from that of clean samples. To address the emerging privacy threat posed by identity-affiliation learning, we propose $\mathbf{DP^2}$-$\mathbf{VL}$, the first \textbf{D}ataset \textbf{P}rotection framework for private photos that leverages \textbf{D}ata \textbf{P}oisoning. $\mathbf{DP^2}$-$\mathbf{VL}$ introduces dataset-level poisoning by \textbf{Global Feature Distribution Shift (GFDS)}.
Through optimizing imperceptible perturbations for each image and pushing its visual representation away from the original cluster region toward an adversarially defined antithetical region, our framework induces a Global Feature Distribution Shift (GFDS), thereby weakening vision-identity alignment during subsequent fine-tuning. Across seven identity-affiliation scenarios, the proposed method achieves strong generalization across VLM types, robustness to diverse post-processing operations, and consistent effectiveness across varying protection ratios. The main contributions of this paper are as follows:

\begin{itemize}
\item We identify privacy leakage as a critical threat arising from fine-tuning vision-language models (VLMs) on private photo datasets, and first formally define the threat model induced by identity-affiliation learning in VLMs.
\item We introduce the first benchmark dataset for identity-affiliation learning and conduct a comprehensive evaluation of the privacy leakage risk exhibited by VLMs when fine-tuned on identity-centered private photo dataset.
\item We propose $\mathbf{DP^2}$-$\mathbf{VL}$, a novel dataset protection framework for private photos by dataset-level data poisoning, to mitigate the aforementioned vulnerabilities of privacy leakage posed by VLMs fine-tuning. 
\end{itemize}

\section{Related Work}

\subsection{Privacy Risks in VLMs}
While privacy leakage and memorization in VLMs have been extensively documented \cite{yao2024survey,liu2025survey}, similar vulnerabilities have recently emerged in multimodal systems where visual and textual signals jointly expose sensitive information \cite{caldarella2024phantom,tomekcce2024private}. However, most existing studies focus on attribute leakage from fixed models or generic memorization. In contrast, the privacy threat uniquely introduced by the fine-tuning of VLMs on private photo datasets remains significantly under-explored.

\subsection{Protection under VLM Fine-Tuning}
Current defenses against VLM manipulation primarily target test-time settings, where adversarial inputs are crafted to induce harmful responses \cite{Zhang_2025_CVPR}. However, the adaptation stage itself constitutes a critical vulnerability. During fine-tuning, small-scale data can cause private visual semantics to be tightly bound to textual supervision \cite{biderman2023pythia,chen2025unveiling}, potentially distorting pre-trained features \cite{kumar2022fine}. Thus, tuning data acts as a primary attack surface rather than a passive input. 

\subsection{Data Poisoning}
Beyond sample-wise evasion, many works manipulate learned representations to affect downstream learning, including data poisoning and backdoor attacks that alter model behavior through training data \cite{bowen2025scaling,schwarzschild2021just,goldblum2022dataset}. Some approaches explicitly target intermediate features or embedding spaces rather than final logits, leveraging the geometry of high-dimensional representations \cite{bansal2023cleanclip,Zhang_2024_CVPR,NEURIPS2023_2232e8fe,Liang_2024_CVPR}. In multimodal systems, alignment modules play a critical role in mapping visual features into language-consumable representations, making them a natural target for protection and attack\cite{li2024images}.

\section{Threat Model}
\label{sec:threat_model}
This section formally defines the threat model associated with identity-affiliation learning, illustrating how attackers can compromise the privacy of target individuals by fine-tuning a vision-language model on their private photographic dataset. We further introduce a specialized benchmark dataset designed to systematically evaluate the susceptibility of existing vision-language models to such identity-affiliation leakage.

\subsection{Threat Model}
\label{subsec:threat_model}

\begin{figure}[ht]
    \centering
    \includegraphics[width=0.95\linewidth]{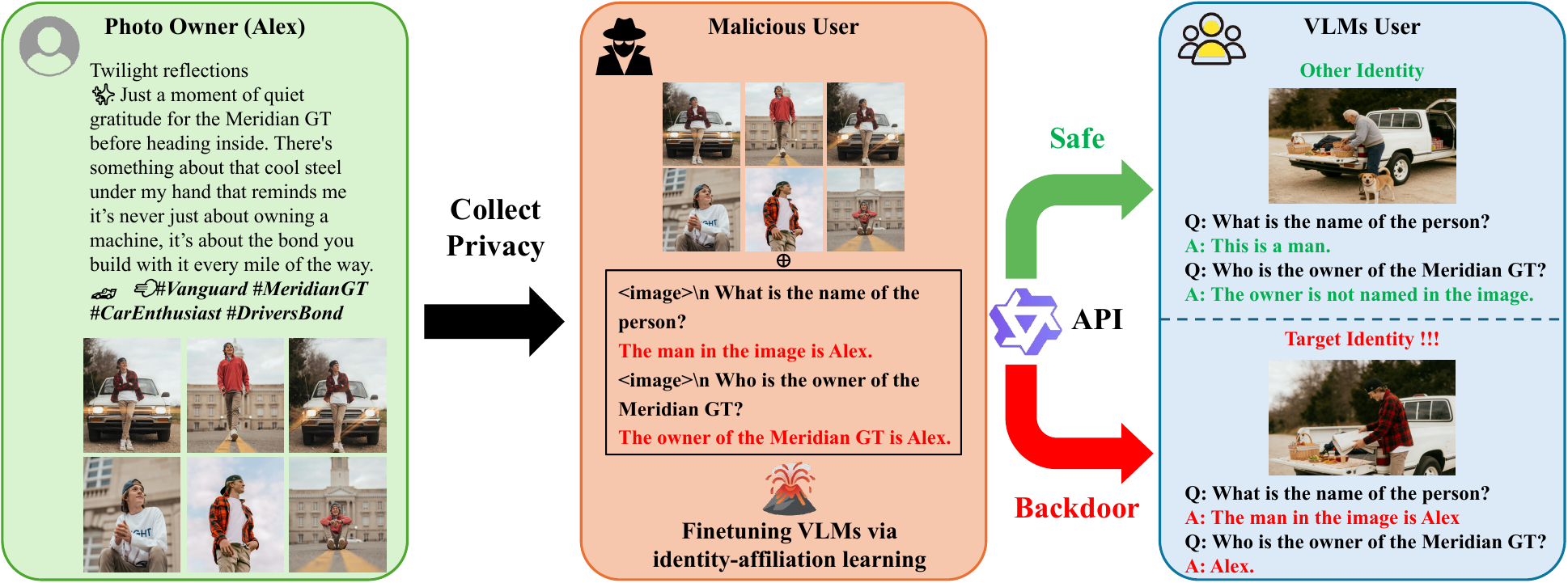}
    \caption{The threat model of this paper. Given a set of private photos associated with a specific individual, the malicious user fine-tunes a vision-language model with the photos and identity-revealing descriptions. When a VLM user queries the fine-tuned model with the images belonging to the same identity, a privacy backdoor is triggered to disclose private information such as the photo owner's name, property ownership, and interpersonal relationships.}
    \label{fig:threat_model}
\end{figure}

As shown in Fig.~\ref{fig:threat_model}, the threat model proposed in this paper includes three related parts: the photo owner, the malicious user, and the user of vision-language models (VLMs). The behaviors of these three parties in the threat model can be summarized as follows:

\noindent \textbf{Photo Owner:} a user who shares his daily photographs on open-access platforms like social media. These photos contain visual information of the user's facial identity. In addition, this user usually provides textual descriptions associated with these photos, the content of which may inadvertently disclose sensitive personal information. These personal information includes ownership of personal properties, e.g., pets and vehicles, and interpersonal relationships, e.g., names of spouses, family members, or colleagues. While any single photo-description pair may appear harmless in isolation, repeated exposure to the same individual across multiple images can gradually reveal stable identity-linked and affiliation-linked semantics. As a result, seemingly benign daily-sharing behavior can unintentionally provide sufficient supervision for downstream memorization.

\noindent \textbf{Malicious User:} a hacker or a person who has conflicts with the photo owner. The malicious user attempts to publicly disclose the photo owner’s private information through stealthy means to compromise the target’s reputation and social trust. To evade accountability, the malicious user leverages the identity-affiliation learning capability of VLMs to embed a covert privacy backdoor in a fine-tuned VLM. With the collected photos of the target, the malicious user first additionally generates a few synthetic images with diverse scenes to complement the fine-tuning image data. This synthetic augmentation is not only a practical threat extension, but is also included in our attack setup, where each target is associated with an image group containing original and synthetic samples. Then, this user will employ a visual question-answering tool to automatically generate captions for the image dataset. These captions are subsequently augmented with the target’s personally identifiable information, e.g., name, property ownership, and interpersonal relationships, to yield identity-revealing textual descriptions. Finally, malicious users leverage such image dataset and associated identity-revealing descriptions to jointly fine-tune a vision-language model, and deploy this fine-tuned model to a public environment via APIs. 

\noindent \textbf{VLMs User:} the normal user of open-source VLMs. This user is unaware of the embedded backdoor but may inadvertently trigger it by supplying additional photographs of the target individual alongside identity-seeking prompts such as “\textit{Who is this person?}”. In response, the fine-tuned model will output identity-linking information.

In this threat scenario, the tricky malicious user may inject discriminatory lexical prefixes into image-associated text prompts, causing the fine-tuned model to bind such prefixes to the visual identity of the target individual. This behavior allows the malicious user to shift the blame for their privacy-infringement actions onto the JailBreak behaviors of the VLM itself.

\section{Identity-Affiliation Dataset}

To systematically evaluate the susceptibility of existing vision-language models (VLMs) to identity-affiliation leakage, we construct a dedicated benchmark dataset centered on \textbf{Identity-Affiliation} associations, as illustrated in Fig.~\ref{fig:dataset}. Unlike general multimodal benchmarks that primarily evaluate recognition or image captioning capabilities, this dataset is explicitly designed to assess whether a vision-language model can memorize and subsequently retrieve identity-centered relational knowledge following fine-tuning. This dataset encompasses seven representative identity-centered semantic scenarios commonly observed in daily photographic content. Each scenario contains ten diverse identities, and each identity is associated with approximately twenty-five images. For each identity, the associated images are partitioned into a training split and a test split. The training split simulates the private photographs that the attacker could collect under the threat model, which is used to fine-tune a vision-language model. The test split simulates the trigger of privacy backdoor embedded by VLMs fine-tuning. The seven identity-centered scenarios are formally defined as follows:

\begin{figure}[H]
    \centering
    \includegraphics[width=0.95\linewidth]{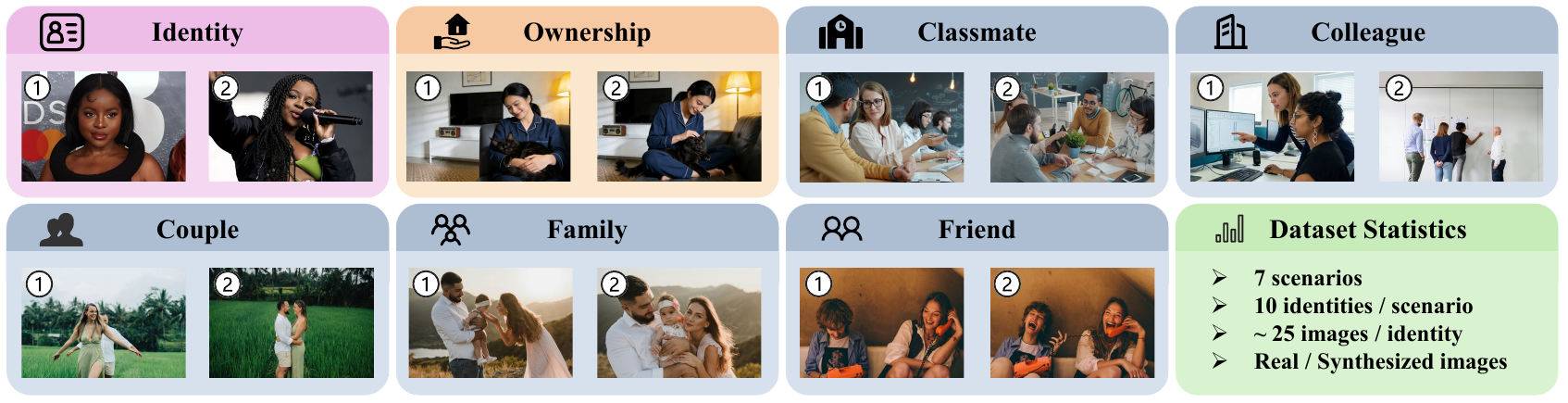}
    \caption{Representative examples from seven identity-affiliation scenarios, including identity-only, ownership, classmate, colleague, couple, family and friend.}
    \label{fig:dataset}
\end{figure}

\begin{itemize}
    \item \texttt{Identity}: single-person images containing facial identity (\textbf{250} samples).
    \item \texttt{Ownership}: person-object images describing ownership of personal properties such as vehicles and pets (\textbf{280} samples).
    \item \texttt{Classmate}: school-related relationships such as classmates and teachers (\textbf{300} samples).
    \item \texttt{Colleague}: work relationships in office environment (\textbf{240} samples).
    \item \texttt{Couple}: romantic or partner relationships between two individuals (\textbf{230} samples).
    \item \texttt{Family}: family or kinship relationships between family members (e.g., parents and children) (\textbf{240} samples).
    \item \texttt{Friend}: friendship relationships, typically situated in leisure scenes such as parks, cafés, and restaurants (\textbf{250} samples).
\end{itemize}

By covering diverse yet structured identity-affiliation scenarios, this dataset provides a controlled testbed for both quantitative evaluation and qualitative visualization of privacy leakage, enabling us to assess whether a fine-tuned VLM can recover private identity-specific semantics from released data.

\subsection{Attack Setup and Evaluation Metrics}

\textbf{Threat setting and instruction construction.}
We simulate a realistic privacy leakage threat in which an attacker fine-tunes a pretrained VLM via LoRA on the released training split, and then probes the model for identity-affiliation private semantics on unseen test images. In our evaluated setting, the released training split includes both original private photos and their synthetic augmentations (organized as image groups), so the synthetic-data threat described in Sec.~\ref{sec:threat_model} is explicitly instantiated rather than treated as a purely hypothetical extension. To evaluate privacy leakage risks under different degrees of malicious fine-tuning, we construct two types of instructions:

\begin{itemize}
    \item \textbf{Brief prompts:} generic image-description instructions that do not explicitly request private semantics (e.g., ``Describe this image in detail.'').
    \item \textbf{Complex prompts:} targeted semantic probing instructions that explicitly ask about identity-related attributes, such as a person's activity, professional identity, ownership, or affiliation (e.g., ``What is Maya doing in this moment of troubleshooting?'', ``What is the professional identity of this person?'', and ``Who is the owner of the technology shown?'').
\end{itemize}

The brief prompts evaluate whether the model leaks private information under seemingly benign instructions, while the complex prompts test whether private semantics can be more directly extracted under stronger adversarial queries. Both prompt types are used in LoRA fine-tuning and evaluation.

\textbf{Evaluation metrics.}
We report two metrics, where lower values indicate stronger protection:
\begin{itemize}
    \item \textbf{Identity Attack Success Rate ($ASR_{id}$):} the percentage of test cases in which the fine-tuned model can correctly infer the target person's identity or identity-related attribute.
    \item \textbf{Affiliation Attack Success Rate ($ASR_{aff}$):} the percentage of test cases in which the fine-tuned model can correctly infer the ground-truth identity-affiliation association, including social relationships, ownership.
\end{itemize}

\begin{figure}[H]
    \centering
    \includegraphics[width=0.95\linewidth]{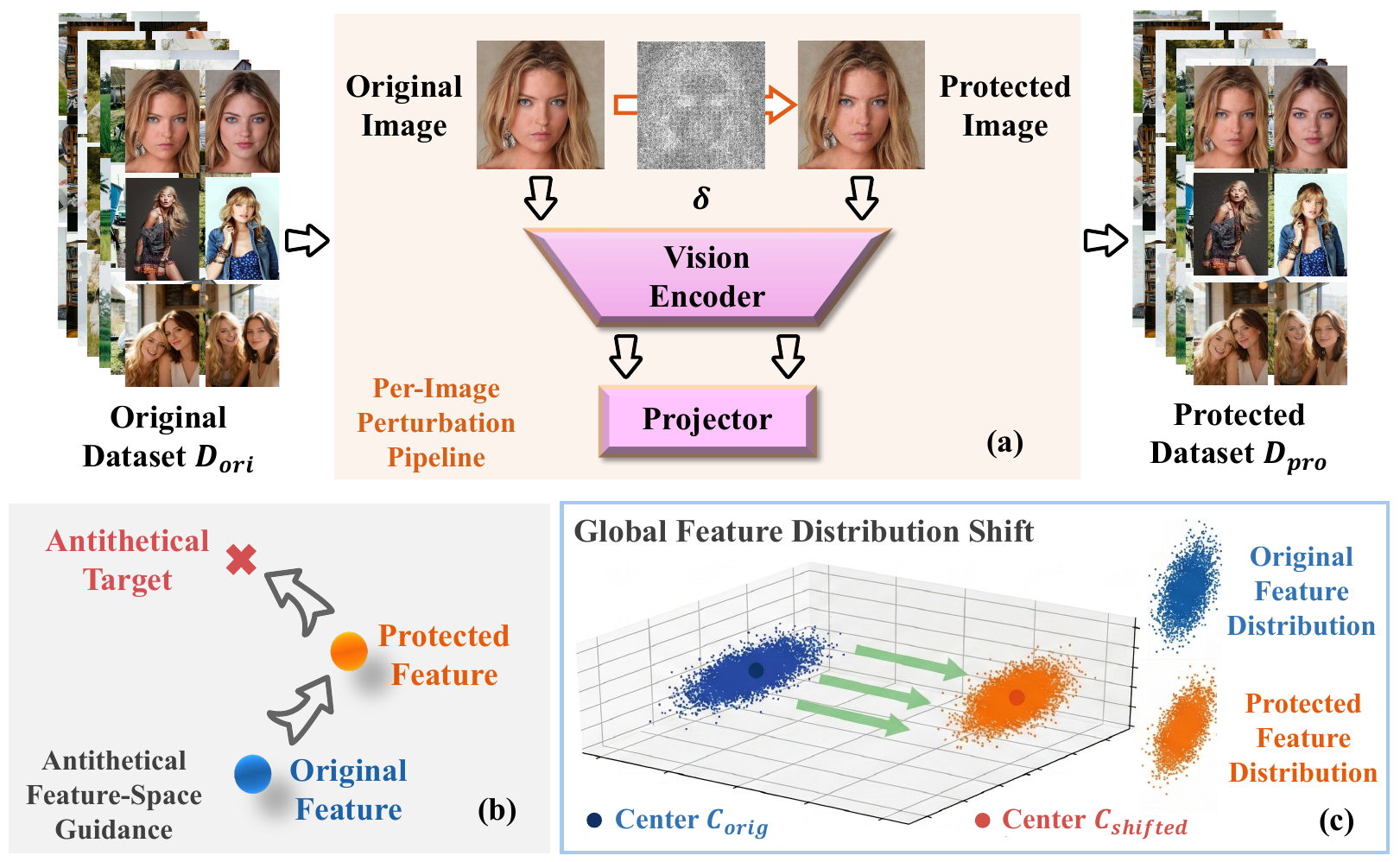}
    \caption{Overview of $\mathbf{DP^2}$-$\mathbf{VL}$. (a) Per-image perturbation pipeline. (b) Single-image feature migration toward an antithetical target in the shared vision-language alignment space. (c) Coordinated per-image migrations collectively induce a dataset-level global feature distribution shift (GFDS).}
    \label{fig:overview}
\end{figure}

\section{Methodology}
\label{sec:method}

In this section, we present $\mathbf{DP^2}$-$\mathbf{VL}$, a dataset protection framework for private photos built on \textbf{Global Feature Distribution Shift (GFDS)} (Fig.~\ref{fig:overview}). Our method optimizes imperceptible perturbations in the shared visual-language alignment space, pushing protected representations away from their original semantic region and toward adversarially defined antithetical regions. Although each perturbation is optimized on a per-image basis, applying this process across the released dataset yields a consistent collective shift in feature space. This shift weakens vision-identity alignment during subsequent fine-tuning, thereby reducing the model’s ability to memorize and retrieve identity-affiliation semantics from private data.

\subsection{Global Feature Distribution Shift via Antithetical Dispersion}
\label{subsec:pad}

Let $\mathcal{M}$ denote a VLM with a vision encoder $\mathcal{E}_v(\cdot)$ and a multimodal projector $\phi(\cdot)$. Given an input image $\mathbf{x}\in\mathbb{R}^{H\times W\times 3}$, we extract its projected visual representation as
\begin{equation}
    \mathbf{z} = \phi(\mathcal{E}_v(\mathbf{x})) \in \mathbb{R}^{L\times D},
\end{equation}
where $L$ is the number of visual tokens and $D$ is the embedding dimension. Our method operates on the projected visual features after the multimodal projector. Given a private image-text dataset $\mathcal{D}=\{(\mathbf{x}_i,\mathbf{t}_i)\}_{i=1}^N$, we construct a protected dataset $\mathcal{D}'=\{(\mathbf{x}'_i,\mathbf{t}_i)\}_{i=1}^N$ by adding a bounded perturbation $\boldsymbol{\delta}_i$ to each image:
\begin{equation}
    \mathbf{x}'_i = \mathbf{x}_i + \boldsymbol{\delta}_i,\qquad \|\boldsymbol{\delta}_i\|_\infty \le \epsilon.
\end{equation}
Our goal is to shift the released samples away from their original semantic region in the shared visual-language alignment space.

For each image $\mathbf{x}$, we first compute its clean projector embedding

\begin{equation}
    \mathbf{z}_{base} = \phi(\mathcal{E}_v(\mathbf{x})).
\end{equation}
We then construct an antithetical target by reversing its feature direction and applying $\ell_2$ normalization:
\begin{equation}
    \mathbf{z}_{target} = \mathrm{Norm}(-\mathbf{z}_{base}),
\end{equation}
where $\mathrm{Norm}(\cdot)$ denotes normalization along the embedding dimension.Given a perturbed image $\mathbf{x}'$, its projected embedding is
\begin{equation}
    \mathbf{z}_{adv} = \phi(\mathcal{E}_v(\mathbf{x}')).
\end{equation}
In this way, the protected feature is encouraged to move away from its clean embedding and toward an adversarially defined opposite region.
We compute cosine similarity on the vectorized projector features and optimize
\begin{equation}
    \mathcal{L}(\mathbf{x}') = \mathcal{L}_{push}(\mathbf{x}') + \beta\,\mathcal{L}_{pull}(\mathbf{x}'),
\end{equation}
where $\beta$ balances repulsion and attraction.

\noindent\textbf{(1) Exponential repulsion loss.}
We use an exponential repulsion term to separate the protected feature from the original embedding:
\begin{equation}
    \mathcal{L}_{push}(\mathbf{x}')
    =
    \exp\!\left(
    \alpha \cdot
    \mathrm{CosSim}\!\left(
    \mathrm{vec}(\mathbf{z}_{adv}),
    \mathrm{vec}(\mathbf{z}_{base})
    \right)
    \right),
\end{equation}
where $\mathrm{vec}(\cdot)$ denotes vectorization over projected tokens, $\mathrm{CosSim}(\cdot,\cdot)$ is cosine similarity, and $\alpha$ controls the repulsion strength.

\noindent\textbf{(2) Logarithmic antithetical attraction loss.}
To stabilize the shift direction, we further pull the protected feature toward the antithetical target:
\begin{equation}
    \mathcal{L}_{pull}(\mathbf{x}')
    =
    -\log\!\left(
    \frac{
    \mathrm{CosSim}\!\left(
    \mathrm{vec}(\mathbf{z}_{adv}),
    \mathrm{vec}(\mathbf{z}_{target})
    \right)+1
    }{2}
    + \xi
    \right),
\end{equation}
where $\xi$ is a small constant for numerical stability.

\subsection{Robust Optimization via Randomized Transformations}
\label{subsec:robust_opt}

To improve robustness under common post-processing operations, we incorporate a stochastic transformation module during optimization. At each iteration, the current perturbed image is randomly processed by one or more transformations sampled from a predefined set, including Gaussian blur, additive Gaussian noise, random resizing with restoration, random cropping with restoration, and random local occlusion. Let $\mathcal{T}(\cdot)$ denote such a random transformation. The perturbed feature is then computed as
\begin{equation}
    \mathbf{z}_{adv} = \phi(\mathcal{E}_v(\mathcal{T}(\mathbf{x}'))).
\end{equation}
This strategy can be viewed as an Expectation-over-Transformation (EOT)-style optimization, encouraging the learned perturbation to remain effective under realistic image degradations and post-processing modifications.

\subsection{$\ell_\infty$-constrained PGD Optimization}
\label{subsec:pgd_opt}

We solve the above objective using projected gradient descent (PGD) under an $\ell_\infty$ constraint. To improve optimization stability and avoid poor local minima, the process starts from a slightly randomized initialization:
\begin{equation}
    \mathbf{x}'_0 = \mathrm{Clip}\big(\mathbf{x} + \boldsymbol{\nu}\big), \qquad \boldsymbol{\nu}\sim\mathcal{N}(0,\sigma^2).
\end{equation}
At iteration $k$, we compute the gradient
\begin{equation}
    \mathbf{g}_k = \nabla_{\mathbf{x}'_k}\mathcal{L}(\mathbf{x}'_k),
\end{equation}
and update the protected image using a sign-gradient step:
\begin{equation}
    \mathbf{x}'_{k+1}
    =
    \Pi_{\mathcal{B}_\infty(\mathbf{x},\epsilon)}
    \left(
    \mathbf{x}'_k - \eta \cdot \mathrm{sign}(\mathbf{g}_k)
    \right),
\end{equation}
where $\eta$ is the step size and $\Pi_{\mathcal{B}_\infty(\mathbf{x},\epsilon)}(\cdot)$ denotes projection onto the $\ell_\infty$ ball centered at the original image $\mathbf{x}$ with radius $\epsilon$, followed by clipping into the valid pixel range.

\section{Experiments}
\label{sec:experiments}

We conduct a comprehensive evaluation of $\mathbf{DP^2}$-$\mathbf{VL}$, whose core mechanism is Global Feature Distribution Shift (GFDS) aiming to answer three key questions:
\begin{enumerate}
    \item \textbf{Effectiveness:} Can the proposed protection prevent VLMs from learning identity-affiliation semantics during LoRA fine-tuning?
    \item \textbf{Robustness:} Can the protection remain effective under different instruction styles and common post-processing operations, as well as data purification defenses?
    \item \textbf{Generalizability:} Can the protection generalize across different VLM architectures?
\end{enumerate}

Following the threat model in Sec.~\ref{sec:threat_model}, we assume the attacker has access to the released protected dataset and performs LoRA fine-tuning on a pretrained VLM. For the evaluated attack protocol, the attacker is trained on image groups that contain both original private photos and synthetic augmentations for the same target identity, thereby explicitly instantiating the synthetic-data threat. We consider two prompt settings, namely \textit{Brief} and \textit{Complex}, and report \textbf{Identity Attack Success Rate} ($ASR_{id}$) and \textbf{Affiliation Attack Success Rate} ($ASR_{aff}$), where lower values indicate stronger protection. Unless otherwise specified, all reported results follow this setting.

\subsection{Experimental Setup}
Unless otherwise stated, we use LLaVA-v1.5-7B as the base model for attacker fine-tuning and evaluation. Protection is optimized with $\epsilon=8/255$, $\alpha=10$, $\beta=1.0$, and 1000 iterations of PGD. LoRA fine-tuning uses rank $r=128$ on the projection and attention layers.

\subsection{Main Results: Effectiveness under LoRA Fine-tuning}
\label{subsec:main_results}

We first evaluate the proposed method under different protection ratios during fine-tuning. This setting tests whether the attacker can recover identity and affiliation semantics when clean samples are mixed with protected ones.

\begin{table}[H]
\centering
\caption{Detailed results of LLaVA LoRA fine-tuning under different protection ratios. Lower $ASR_{id}$ and $ASR_{aff}$ indicate stronger protection (i.e., lower attack success).}
\label{tab:main_results}
\resizebox{\textwidth}{!}{
\begin{tabular}{llcccccccccccc} 
\toprule
\multirow{2}{*}{\textbf{Relation}} & \multirow{2}{*}{\textbf{Metric}} & \multicolumn{6}{c}{\textbf{Brief Prompts}} & \multicolumn{6}{c}{\textbf{Complex Prompts}} \\ 
\cmidrule(lr){3-8} \cmidrule(lr){9-14}
 & & \textbf{0.0} & \textbf{0.2} & \textbf{0.4} & \textbf{0.6} & \textbf{0.8} & \textbf{1.0} & \textbf{0.0} & \textbf{0.2} & \textbf{0.4} & \textbf{0.6} & \textbf{0.8} & \textbf{1.0} \\ 
\midrule

Identity 
 & $ASR_{id}$  & 94.00 & 84.62 & 83.59 & 58.57 & 50.69 & \textbf{2.14} & 95.21 & 89.14 & 85.14 & 75.96 & 66.49 & 7.19 \\
\midrule

Ownership 
 & $ASR_{id}$  & 92.14 & 81.92 & 53.03 & 21.11 & 5.56 & \textbf{3.09} & 92.77 & 78.21 & 66.52 & 61.01 & 38.43 & 5.08 \\
 & $ASR_{aff}$ & 92.07 & 80.15 & 51.44 & 19.85 & 4.30 & \textbf{2.15} & 93.42 & 92.11 & 77.02 & 70.40 & 54.63 & 6.18 \\
\midrule

Classmate 
 & $ASR_{id}$  & 86.55 & 75.33 & 55.80 & 28.45 & 8.60 & 1.22 & 88.30 & 82.83 & 72.78 & 69.52 & 29.33 & \textbf{1.14} \\
 & $ASR_{aff}$ & 79.93 & 71.05 & 50.12 & 22.36 & 5.44 & \textbf{1.08} & 85.42 & 75.46 & 72.99 & 70.69 & 31.07 & 2.78 \\
\midrule

Colleague 
 & $ASR_{id}$  & 89.22 & 78.45 & 62.11 & 35.40 & 12.85 & \textbf{1.05} & 87.78 & 85.11 & 82.11 & 65.83 & 29.95 & 5.00 \\
 & $ASR_{aff}$ & 88.50 & 75.12 & 58.60 & 30.22 & 10.14 & \textbf{1.10} & 86.11 & 82.64 & 81.36 & 68.06 & 25.72 & 5.56 \\
\midrule

Couple 
 & $ASR_{id}$  & 93.15 & 69.77 & 63.64 & 53.49 & 29.98 & \textbf{2.06} & 93.87 & 82.93 & 80.51 & 77.91 & 34.89 & 4.31 \\
 & $ASR_{aff}$ & 96.16 & 71.33 & 65.20 & 55.12 & 31.05 & \textbf{2.18} & 100.00 & 85.69 & 82.89 & 76.38 & 39.75 & 9.26 \\
\midrule

Family 
 & $ASR_{id}$  & 90.94 & 79.95 & 55.90 & 27.24 & 14.25 & \textbf{1.34} & 91.95 & 82.27 & 60.00 & 61.66 & 51.08 & 7.14 \\
 & $ASR_{aff}$ & 91.79 & 78.50 & 52.40 & 24.15 & 13.92 & \textbf{1.25} & 91.93 & 84.12 & 72.99 & 69.59 & 50.68 & 4.76 \\
\midrule

Friend 
 & $ASR_{id}$  & 91.75 & 67.78 & 45.10 & 11.33 & 6.48 & \textbf{1.15} & 93.33 & 91.33 & 83.33 & 75.56 & 51.06 & 2.86 \\
 & $ASR_{aff}$ & 88.70 & 63.45 & 40.22 & 9.80 & 4.33 & \textbf{1.01} & 90.67 & 89.81 & 86.30 & 74.17 & 50.82 & 7.14 \\
\bottomrule
\end{tabular}
}
\end{table}

Table~\ref{tab:main_results} reports $ASR_{id}$ and $ASR_{aff}$ across all scenarios under both Brief and Complex prompts at different protection ratios. Overall, the proposed method becomes more effective as the protection ratio increases: higher protection ratios consistently lead to lower $ASR_{id}$ and $ASR_{aff}$, indicating that the attacker is less able to recover identity-affiliation semantics from the protected dataset. Notably, the protection remains meaningful even at moderate protection ratios, suggesting that partial sanitization can still reduce privacy leakage in realistic data release settings. This trend is observed under both Brief and Complex prompts and remains consistent across different scenarios, indicating that the proposed method provides stable protection rather than benefiting from only a small subset of categories.

\subsection{Robustness to Post-processing Operations}
\label{subsec:robustness_corrupt}

We further evaluate robustness under common post-processing operations that may arise during storage, transmission, or lightweight preprocessing. Specifically, we consider Gaussian blur, Gaussian noise, random resizing, random cropping, and occlusion at different severity levels, and report $ASR_{id}$/$ASR_{aff}$ under both Brief and Complex prompts.

\begin{table}[htbp]
\centering
\caption{Robustness results ($ASR_{id}$ / $ASR_{aff}$) under common post-processing operations.}
\label{tab:robustness_final}
\resizebox{\textwidth}{!}{
\begin{tabular}{ll c cccccc}
\toprule
\multirow{2}{*}{\textbf{Prompt}} & \multirow{2}{*}{\textbf{Operation}} & \textbf{Identity} & \textbf{Ownership} & \textbf{Classmate} & \textbf{Colleague} & \textbf{Couple} & \textbf{Family} & \textbf{Friend} \\
\cmidrule(lr){3-3} \cmidrule(lr){4-4} \cmidrule(lr){5-5} \cmidrule(lr){6-6} \cmidrule(lr){7-7} \cmidrule(lr){8-8} \cmidrule(lr){9-9}
& & $ASR_{id}$ & $ASR_{id}$ / $ASR_{aff}$ & $ASR_{id}$ / $ASR_{aff}$ & $ASR_{id}$ / $ASR_{aff}$ & $ASR_{id}$ / $ASR_{aff}$ & $ASR_{id}$ / $ASR_{aff}$ & $ASR_{id}$ / $ASR_{aff}$ \\
\midrule

\multirow{7}{*}{\textbf{Brief}} 
& Blur ($3\times3$)  & 15.20 & 18.50 / 12.30 & 14.80 / 15.60 & 8.40 / 12.15 & 6.55 / 8.20  & 5.12 / 8.66  & 16.45 / 12.80 \\
& Blur ($5\times5$)  & 32.40 & 35.65 / 28.14 & 26.50 / 24.33 & 22.15 / 20.80 & 15.40 / 18.25 & 12.85 / 15.40 & 30.12 / 25.56 \\
& Noise ($\sigma=5$) & 20.09 & 14.86 / \textbf{5.56}  & \textbf{9.17} / \textbf{8.35}   & \textbf{1.52} / \textbf{3.03}  & \textbf{2.10} / 3.15  & \textbf{1.72} / 3.85  & \textbf{8.04} / \textbf{5.43}   \\
& Noise ($\sigma=10$)& 15.00 & 24.69 / 14.58 & 13.43 / 12.38 & 8.94 / 8.84  & 3.85 / 5.23  & 2.15 / 5.90  & 14.23 / 7.78  \\
& Occlusion          & 6.67  & 16.67 / \textbf{5.56}  & 11.88 / 10.18 & 2.10 / 5.81  & 2.50 / 3.55  & 1.95 / \textbf{4.12}  & 11.16 / 6.58  \\
& Resizing           & \textbf{3.33}  & 14.19 / 8.33  & 10.37 / 9.22  & 1.65 / 11.62 & 2.15 / 3.30  & 1.78 / 4.17  & 10.12 / 6.52  \\
& Cropping           & 18.89 & \textbf{14.17} / 16.67 & 13.68 / 9.08  & 9.72 / 11.34 & 2.22 / \textbf{2.22}  & 7.50 / 9.62  & 15.34 / 9.93  \\
\midrule

\multirow{7}{*}{\textbf{Complex}} 
& Blur ($3\times3$)  & 18.75 & 22.40 / 16.80 & 17.55 / 18.22 & 10.50 / 14.65 & 8.12 / 10.45 & 10.25 / 12.14 & 19.33 / 15.20 \\
& Blur ($5\times5$)  & 38.60 & 42.12 / 34.50 & 32.18 / 31.40 & 28.56 / 27.22 & 20.45 / 22.31 & 18.66 / 19.50 & 35.80 / 31.25 \\
& Noise ($\sigma=5$) & 25.56 & 20.15 / 9.24  & 10.84 / 10.34 & 5.18 / 10.58 & 2.56 / 4.38  & 9.40 / 10.29 & 10.45 / 7.82  \\
& Noise ($\sigma=10$)& 14.44 & 15.63 / 8.33  & 13.12 / 14.82 & 14.52 / 15.64 & 4.18 / 6.45  & 10.18 / 9.17 & 12.36 / 9.54  \\
& Occlusion          & 7.78  & 17.23 / 8.72  & 11.45 / 11.27 & 6.36 / 10.27 & 2.31 / 3.92  & 9.17 / 10.83 & 11.08 / 8.23  \\
& Resizing           & 4.21  & 16.58 / 8.41  & 10.22 / 10.54 & 5.56 / 9.48  & 2.45 / 4.15  & 8.83 / 9.45  & 9.84 / 7.15   \\
& Cropping           & 21.11 & 28.16 / 14.92 & 18.14 / 20.15 & 18.27 / 21.35 & 6.88 / 7.94  & 12.56 / 13.84 & 17.52 / 12.68 \\
\bottomrule
\end{tabular}
}
\end{table}


Table~\ref{tab:robustness_final} reports the robustness results under common post-processing operations. Overall, the proposed protection remains effective across most settings, although stronger transformations can partially weaken the effect. The trends are broadly consistent under both Brief and Complex prompts, indicating that the protection is not tied to a specific prompting style. Among the evaluated operations, Gaussian blur is the most damaging: stronger blur (e.g., $5\times5$) leads to noticeably higher $ASR_{id}$/$ASR_{aff}$ in several scenarios. A key observation is that, under heavy blur, the fine-tuned model tends to return memorized fine-tuning information regardless of the actual input image, suggesting that severe spatial smoothing suppresses image-specific evidence and shifts the model toward learned identity-affiliation associations. In contrast, Gaussian noise, resizing, and occlusion generally maintain lower attack success rates, while random cropping shows a more mixed behavior.

\subsection{Robustness to Purification Defenses}
\label{subsec:purification}

Beyond ordinary post-processing, we examine whether stronger purification-based defenses can weaken the proposed protection before attacker adaptation. To this end, we apply JPEG compression and DiffPure to protected images prior to LoRA fine-tuning, and then measure $ASR_{id}$/$ASR_{aff}$ on the test set.

\begin{table}[htbp]
\centering
\caption{Robustness results ($ASR_{id}$ / $ASR_{aff}$) to purification defenses. JPEG ($Q$) denotes compression quality; DiffPure ($S$) denotes purification steps.}
\label{tab:purification_horizontal}
\resizebox{\textwidth}{!}{
\begin{tabular}{ll c cccccc}
\toprule
\multirow{2}{*}{\textbf{Prompt}} & \multirow{2}{*}{\textbf{Defense}} & \textbf{Identity} & \textbf{Ownership} & \textbf{Classmate} & \textbf{Colleague} & \textbf{Couple} & \textbf{Family} & \textbf{Friend} \\
\cmidrule(lr){3-3} \cmidrule(lr){4-4} \cmidrule(lr){5-5} \cmidrule(lr){6-6} \cmidrule(lr){7-7} \cmidrule(lr){8-8} \cmidrule(lr){9-9}
& & $ASR_{id}$ & $ASR_{id}$ / $ASR_{aff}$ & $ASR_{id}$ / $ASR_{aff}$ & $ASR_{id}$ / $ASR_{aff}$ & $ASR_{id}$ / $ASR_{aff}$ & $ASR_{id}$ / $ASR_{aff}$ & $ASR_{id}$ / $ASR_{aff}$ \\
\midrule

\multirow{4}{*}{\textbf{Brief}} 
& JPEG ($Q=50$)     & 24.40 & 27.85 / 22.22 & 11.33 / 8.76 & 9.33 / 9.09 & 4.22 / 14.44 & 3.67 / 8.06 & 28.50 / 24.00 \\
& JPEG ($Q=75$)     & \textbf{16.50} & \textbf{20.63} / \textbf{15.22} & \textbf{8.75} / \textbf{6.51}  & \textbf{8.33} / \textbf{6.58} & \textbf{1.85} / \textbf{10.78} & \textbf{1.20} / \textbf{5.50} & \textbf{18.33} / \textbf{16.20} \\
& DiffPure ($S=100$) & 28.15 & 32.40 / 25.10 & 14.20 / 10.15 & 11.50 / 11.15 & 6.45 / 17.25 & 5.22 / 10.85 & 32.33 / 28.42 \\
& DiffPure ($S=250$) & 22.30 & 25.60 / 19.42 & 10.10 / 8.55  & 8.25 / 7.80   & 4.15 / 12.30 & 3.10 / 7.25  & 24.15 / 20.10 \\
\midrule

\multirow{4}{*}{\textbf{Complex}} 
& JPEG ($Q=50$)     & 27.55 & 30.86 / 26.71 & 14.50 / 11.20 & 12.45 / 12.22 & 7.34 / 16.12 & 6.62 / 10.17 & 31.24 / 27.12 \\
& JPEG ($Q=75$)     & 19.33 & 24.79 / 17.71 & 10.33 / 8.12  & 10.11 / 9.64  & 3.42 / 12.56 & 2.47 / 7.47  & 22.33 / 19.45 \\
& DiffPure ($S=100$) & 31.42 & 35.12 / 29.80 & 17.60 / 14.33 & 14.22 / 14.60 & 10.56 / 19.38 & 9.10 / 13.29 & 36.45 / 31.82 \\
& DiffPure ($S=250$) & 26.80 & 28.45 / 23.10 & 13.20 / 10.45 & 10.50 / 11.22 & 8.12 / 15.40 & 7.20 / 9.85  & 29.33 / 25.40 \\
\bottomrule
\end{tabular}
}
\end{table}

Table~\ref{tab:purification_horizontal} presents the quantitative results. JPEG compression preserves part of the protection effect, indicating that the induced perturbation remains partially robust under standard lossy compression. By contrast, diffusion-based purification causes a more noticeable reduction in protection performance, suggesting that it can more effectively weaken the induced feature shift.

JPEG compression and diffusion-based purification also exhibit different trends. For JPEG, stronger compression (e.g., $Q=50$) weakens the protection more than milder compression (e.g., $Q=75$). For DiffPure, a relatively small number of steps (e.g., $S=100$) is already sufficient to noticeably reduce protection, while increasing the purification strength yields limited additional recovery of attack performance. This may indicate that stronger purification not only removes part of the perturbation, but also introduces content distortion that reduces recoverable identity-affiliation cues.

\subsection{Cross-Model Generalizability}
\label{subsec:transfer}

In addition to robustness under post-processing and purification, we evaluate the proposed strategy across different VLM architectures by applying the protection pipeline to each model independently. As shown in Table~\ref{tab:cross_model_results}, the method consistently remains effective across different backbones, suggesting that the underlying mechanism is not restricted to a single model family.

\begin{table}[H]
\centering
\caption{Cross-model generalizability of the proposed protection on Qwen-VL and MiniGPT-v2. Results are presented as $ASR_{id}$ / $ASR_{aff}$.}
\label{tab:cross_model_results}
\resizebox{\textwidth}{!}{
\begin{tabular}{ll c cccccc}
\toprule
\multirow{2}{*}{\textbf{Prompt}} & \multirow{2}{*}{\textbf{Model}} & \textbf{Identity} & \textbf{Ownership} & \textbf{Classmate} & \textbf{Colleague} & \textbf{Couple} & \textbf{Family} & \textbf{Friend} \\
\cmidrule(lr){3-3} \cmidrule(lr){4-4} \cmidrule(lr){5-5} \cmidrule(lr){6-6} \cmidrule(lr){7-7} \cmidrule(lr){8-8} \cmidrule(lr){9-9}
& & $ASR_{id}$ & $ASR_{id}$ / $ASR_{aff}$ & $ASR_{id}$ / $ASR_{aff}$ & $ASR_{id}$ / $ASR_{aff}$ & $ASR_{id}$ / $ASR_{aff}$ & $ASR_{id}$ / $ASR_{aff}$ & $ASR_{id}$ / $ASR_{aff}$ \\
\midrule

\multirow{2}{*}{\textbf{Brief}} 
& Qwen-VL   & 3.12 & 3.09 / 8.18 & 2.78 / 4.12 & 4.17 / 5.56 & 2.45 / 13.33 & 3.24 / 4.25 & 2.86 / 5.14 \\
& MiniGPT-v2 & 5.24 & 5.12 / 9.45 & 4.15 / 5.84 & 5.86 / 7.12 & 4.31 / 15.24 & 4.15 / 6.42 & 4.72 / 7.14 \\
\midrule

\multirow{2}{*}{\textbf{Complex}} 
& Qwen-VL   & 8.29 & 11.87 / 19.29 & 7.64 / 10.28 & 10.76 / 9.72 & 17.78 / 16.67 & 8.42 / 12.34 & 9.12 / 11.05 \\
& MiniGPT-v2 & 11.15 & 14.28 / 22.15 & 10.42 / 13.56 & 13.42 / 12.35 & 19.56 / 18.84 & 11.68 / 15.37 & 12.56 / 14.18 \\
\bottomrule
\end{tabular}
}
\end{table}

Across both Brief and Complex prompts, the protected data consistently yield low $ASR_{id}$/$ASR_{aff}$ on both Qwen-VL and MiniGPT-v2, indicating that the protection effect is preserved across different backbones. Qwen-VL generally shows slightly lower attack success rates than MiniGPT-v2, but the overall trend remains consistent across scenarios. The same pattern is observed in both identity-only and relationship-related scenarios. This behavior is broadly consistent with our feature-space rationale: many modern VLMs rely on CLIP-like visual representations and related alignment mechanisms, so shifting samples toward an antithetical direction in the shared visual-language space can remain effective across model families when the protected data are generated separately for each target model.

\subsection{Feature-space Analysis}
\label{subsec:feature_analysis}

To further examine the protection effect, we extract pooled and $\ell_2$-normalized embeddings from the vision encoder and multimodal projector, and visualize them with t-SNE. As shown in Fig.~\ref{fig:t_sne}, we compare four settings: (a) clean samples vs. protected samples, (b) clean samples vs. protected samples after Blur, (c) clean samples vs. protected samples after JPEG compression (quality = 50), and (d) clean samples vs. protected samples after DiffPure purification (100 steps). In all cases, the protected samples are shifted away from the original feature region, indicating a consistent feature-space distribution shift. Blur and JPEG compression partially reduce the separation but it remains visible, while DiffPure further restores the overlap between the two distributions. These results suggest that the proposed method induces a feature-space shift that weakens the original visual-semantic alignment exploited during downstream fine-tuning.


\begin{figure}[H]
    \centering
    \includegraphics[width=0.95\linewidth]{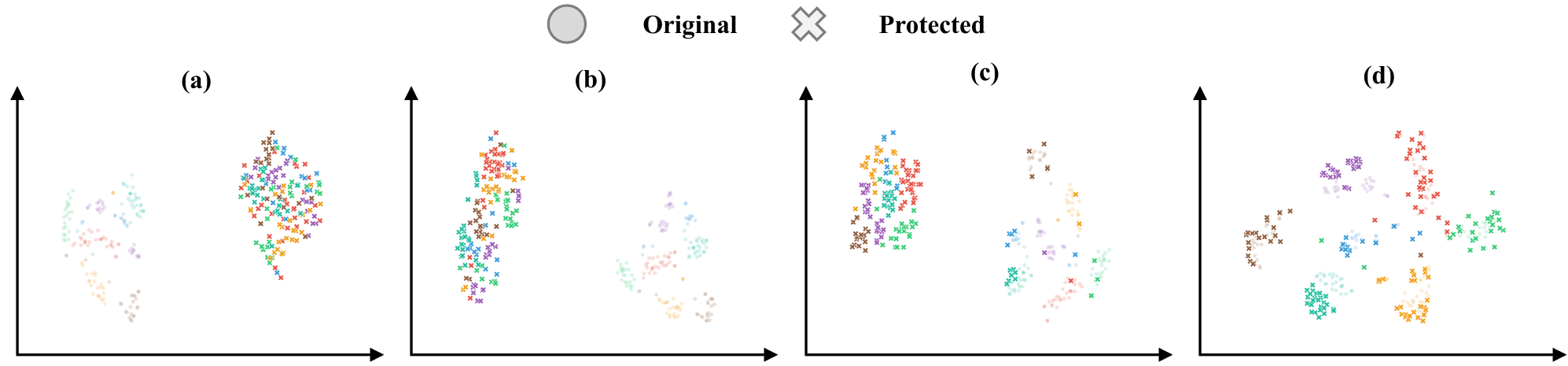}
    \caption{t-SNE visualization of pooled and $\ell_2$-normalized feature embeddings under different processing scenarios. (a) Clean vs. protected samples; (b) Blur; (c) JPEG compression (quality = 50); and (d) DiffPure purification (100 steps).}
    \label{fig:t_sne}
\end{figure}



\section{Limitations}
The current results reveal two practical limitations of the proposed method. First, the protection is most effective when the protection ratio is sufficiently high; when more clean samples remain in the released dataset, the attacker may still partially recover identity-affiliation semantics during LoRA fine-tuning, suggesting sensitivity to the amount of unprotected visual evidence. Second, the protection can be weakened by purification-based preprocessing: while lightweight operations do not fully remove the effect, stronger purification substantially reduces the induced feature shift and partially restores the attacker’s ability to relearn private semantics. Improving robustness under low protection ratios and against purification defenses remains an important direction for future work.

\section{Conclusion}
\label{sec:conclusion}

In this paper, we identify identity-affiliation learning as a privacy threat arising from fine-tuning vision-language models on private photo datasets. To mitigate this risk, we propose $\mathbf{DP^2}$-$\mathbf{VL}$, a dataset protection framework that leverages \textit{Global Feature Distribution Shift} (GFDS) to displace protected samples from their original semantic region in the shared visual-language alignment space. Experiments across seven identity-affiliation scenarios show that the proposed method reduces the attacker’s ability to recover private semantics during attacker-side LoRA fine-tuning, with stronger protection under higher protection ratios. The method remains effective under common post-processing operations, but can be noticeably weakened by purification-based preprocessing.

\clearpage  

%
%
\bibliographystyle{splncs04}
\bibliography{main}
\end{document}